\documentclass{article}
\usepackage[dblblindworkshop, final, nonatbib]{neurips_2026}
\workshoptitle{Diffusion Language Models: Foundations, Efficiency, and Reasoning (DiffuLM)}

\usepackage[utf8]{inputenc}
\usepackage[T1]{fontenc}
\usepackage[hidelinks]{hyperref}
\usepackage{url}
\usepackage{booktabs}
\usepackage{amsfonts}
\usepackage{amsmath,amssymb}
\usepackage{nicefrac}
\usepackage{microtype}
\usepackage{xcolor}
\usepackage{graphicx}
\usepackage{caption}
\usepackage{placeins}
\makeatletter
\renewenvironment{table}{\@float{table}}{\end@float}
\makeatother



\newcommand{\C}{\mathrm{C}}
\newcommand{\given}{\,\vert\,}
\newcommand{\Chat}{\widehat{\C}}

\title{How Much Dependency Can a Marginal Rule See?\\
Total Correlation in Diffusion LM Parallel Decoding}

\author{%
  Dovud Asadov\thanks{Equal contribution.}\\
  New Uzbekistan University\\
  \texttt{d.asadov@newuu.uz}
  \And
  Rustambek Urokov\footnotemark[1]\\
  New Uzbekistan University\\
  \texttt{r.urokov@newuu.uz}
  \And
  Qobiljon Toshnazarov\\
  New Uzbekistan University\\
  \texttt{q.toshnazarov@newuu.uz}
  \AND
  Abduaziz Kayumov\\
  New Uzbekistan University\\
  \texttt{a.kayumov@newuu.uz}
  \And
  Mardon Hazratov\\
  New Uzbekistan University\\
  \texttt{m.hazratov@newuu.uz}
  \And
  Husanboy Mansuraliyev\\
  New Uzbekistan University\\
  \texttt{h.mansuraliyev@newuu.uz}
}

\begin{document}
\maketitle

\begin{abstract}
A masked diffusion language model is fast because it commits several tokens per step.
Each token is sampled from its own marginal, so the dependency between the tokens is
lost, and the size of that loss is the conditional total correlation $\C$ of the
committed set. Most decoders choose the set from per-position marginals, which can bound
$\C$ but cannot determine it. We ask how well they \emph{predict} it. We estimate $\C$
for fixed candidate sets on four masked diffusion LMs from two families. On three of the
four, the best marginal-only predictor is off by about one bit ($0.74$ to $1.17$), $1.8$
to $3.7$ times less than a baseline that knows only the bound, and most of this accuracy
comes from the bound itself. On a subset, our estimate ranks sets like the exact KL cost
of a parallel commit, but sits a median $0.4$ to $1.1$ bits below it on prose and when
the rest of the text is still masked. Dream-instruct is the exception, mainly on prose,
where its conditionals are least self-consistent. Among six common marginal rules,
pooling over the set beats gating on its weakest position in all $24$ paired
comparisons. The measurement needs a validated readout: Dream and LLaDA place a
position's prediction at \emph{different} logit indices, and the wrong index inflates an
inconsistency diagnostic $14$ to $41\times$. We give three checks that measure the
convention. Our results concern predicting $\C$ for a fixed set, not what a
dependency-aware decoder gains.
\end{abstract}

\section{Introduction}
\label{sec:intro}

A masked diffusion language model (dLLM) \cite{llada,dream} starts from a fully masked
sequence and fills it in over a few denoising steps. Its speed comes from committing
several positions in the same step. Which positions to commit together is therefore the
main decision a decoder makes.

Committing together has a cost. Suppose two masked positions should read either
``New York'' or ``Los Angeles'', each with probability one half. The model's prediction
for each position on its own is then a fair coin: ``New'' or ``Los'' at the first
position, ``York'' or ``Angeles'' at the second. A parallel commit samples the two
positions independently, so half of the time it writes ``New Angeles'' or ``Los York''.
Neither marginal is wrong. What is lost is the dependency between the two positions.

This loss has an exact measure. Committing a set $S$ in one step replaces the joint
distribution over $S$ by the product of its marginals, and the Kullback--Leibler cost of
that replacement is the \emph{conditional total correlation}
\begin{equation}
  \C(S \given c) \;=\; \sum_{i \in S} H(Y_i \given c) \;-\; H(Y_S \given c) \;\ge\; 0,
  \label{eq:tc}
\end{equation}
where $c$ is the already-decoded context. $\C$ is the information the positions of $S$
share given $c$. It is zero exactly when they are conditionally independent, it is one
bit in the example above, and it lower-bounds the KL of any factorised commit
\cite{parallelbench,huang}.

Most decoders do not measure $\C$. They choose $S$ with a rule computed from each
position's marginal separately: token confidence \cite{fastdllm,dynamicdllm}, the top-2
margin, the predictive entropy, or cross-step stability \cite{less,stability}. In the example, a marginal rule sees two fair coins, and it would
see the same two fair coins if the positions were independent and $\C$ were zero. So a
marginal rule cannot \emph{determine} $\C$. It can only bound it from above
(\S\ref{sec:worstcase}). This is known: it is stated in \cite{cocommit} and motivates
several decoders \cite{demask,dos}.

But a decoder never meets the worst case. It meets only the joints that a trained model
produces. The open question, and the one we answer, is:

\begin{quote}
\emph{On a real dLLM, how much of the variation in $\C$ is recoverable from the
marginals, and how much do the rules that read them leave behind?}
\end{quote}

Answering it needs two things that the literature has not put together: $\C$ measured
per candidate set on a real checkpoint, and a readout of that checkpoint one can trust.

\paragraph{Reading the marginals is not trivial.} Dream \cite{dream} keeps the shift of
its autoregressive backbone, so its output at index $i$ predicts the token at $i{+}1$,
while LLaDA \cite{llada} predicts each masked token in place. Two of the most widely
used open checkpoints therefore need \emph{opposite} conventions, and no type check
catches the difference. A pipeline that reads Dream at the wrong index still returns
numbers, and they look plausible (\S\ref{sec:bugimpact}).

\paragraph{Contributions.}
\begin{itemize}\setlength{\itemsep}{1pt}\setlength{\parskip}{0pt}\setlength{\topsep}{2pt}
\item \textbf{The marginals predict $\C$ to about one bit.} We estimate $\C$ per
candidate set on four checkpoints from two families and fit the strongest marginal-only
predictor we could. On three of the four its error is $0.74$ to $1.17$ bits, $1.8$ to
$3.7$ times smaller than that of a baseline that knows only the upper bound. Most of the
accuracy comes from the bound itself, because real sets use about half of it
(\S\ref{sec:explains}).
\item \textbf{How a rule pools matters more than which statistic it reads.} We score six
deployed rules against measured $\C$ with paired tests. Averaging over the set beats
gating on its weakest position in all $24$ comparisons, and the advantage the top-2
margin has on synthetic tasks does not carry over (\S\ref{sec:fidelity}).
\item \textbf{What the estimate means when the model is inconsistent.} A dLLM's
conditionals need not agree with any single joint. We give an exact identity for what
our estimate measures in that case and check it against the KL cost of a parallel commit
(\S\ref{sec:meaning}).
\item \textbf{A readout protocol.} Three checks \emph{measure} the logit-position
convention of a checkpoint rather than assuming it, and an ablation shows what the
wrong convention does to the statistics above (\S\ref{sec:bugimpact}).
\end{itemize}
As a reference point we also state the worst case exactly (\S\ref{sec:worstcase}), a
short characterisation built on the factorisation bound of \cite[Thm.~1]{huang} and the
analysis of \cite[\S4]{parallelbench}.

\section{Setup: total correlation and what the marginals certify}
\label{sec:worstcase}

\paragraph{Parallel commits.} At each step a masked dLLM predicts every masked position
from the current sequence $x_t$, and the predictions are independent of each other:
$p_\theta(x_{t-1}\given x_t)=\prod_i p_\theta(x^i_{t-1}\given x_t)$ \cite{mdlm,md4}. A
decoder picks a set $S$ of masked positions and samples each from its own marginal. With
$|S|=1$ this is any-order autoregression and nothing is lost. With $|S|>1$ the commit
loses the $\C(S\given c)$ of Eq.~\eqref{eq:tc}. We write $Y_i$ for the token at position
$i$, $k=|S|$, and $c$ for everything the model can see. This section assumes that the
model's conditionals \emph{compose into one joint}: there is a single joint distribution
over $Y_S$ whose conditionals they are. \S\ref{sec:meaning} treats the
case where they do not.

\paragraph{The bound $W$.} The marginals fix the first term of Eq.~\eqref{eq:tc}, the
summed marginal entropy $B$. Since the joint entropy is non-negative,
\begin{equation}
  0 \;\le\; \C(S\given c) \;\le\; \sum_{i\in S} H(Y_i\given c) \;=:\; B(S\given c).
  \label{eq:budget}
\end{equation}
A tighter bound uses the fact that a joint is never less uncertain than its most
uncertain coordinate, $H(Y_S\given c)\ge\max_i H(Y_i\given c)$. Let $\C_{\max}$ be the
largest $\C$ over all joints that have the given marginals. Then
\begin{equation}
  \C \;\le\; \C_{\max} \;\le\; W \;:=\; B-\max_i H(Y_i\given c) \;\le\; (1-1/k)\,B .
  \label{eq:width}
\end{equation}
We call $W$ \emph{the bound}. It is the largest dependency that the marginal entropies
allow, and it is all that a marginal rule can certify. In particular, the marginals
certify $\C=0$ if and only if at most one position of $S$ is uncertain.

Inside $[0,\C_{\max}]$ the marginals decide nothing: every value is realised by some
joint, and for equal marginals $\C_{\max}=W$ (App.~\ref{app:fibre}).

\paragraph{Same marginals, different $\C$.} A clean example is $n$ fair coins against
$n$ coins conditioned on exactly $n/2$ heads. Every coin is fair in both, so the
marginals are identical and $B=n$ bits. But the first has $\C=0$ and the second has
$\C=n-\log_2\binom{n}{n/2}$, which is $1.415$ bits at $n{=}4$. Every marginal rule must
give the two the same score \cite{marginalsmatch}.

\paragraph{The six marginal rules.} We score six rules that decoders use. Each reads
one statistic per position and pools it over the set into a \emph{safety score}, where
higher means safer to commit. The statistics are the \emph{confidence} (the largest
probability), the \emph{top-2 margin} (the largest probability minus the second
largest), and the negative \emph{entropy}. Each is pooled in two ways: by the weakest
position of the set (the minimum confidence, the minimum margin, the negative maximum
entropy), or over the whole set (the mean confidence, the mean margin, the negative
summed entropy, which is $-B$). We call the first kind \emph{min} rules and the second
kind \emph{pooled} rules.

\paragraph{Two regimes.} A candidate set is always a block of $k$ adjacent positions,
with $k{=}4$ unless stated. We measure it in two regimes. In the \emph{suffix} regime
everything from the block to the end of the text is masked, which is what a decoder
continuing a prompt sees. In the \emph{infill} regime only the block is masked, with
context on both sides. Suffix blocks are much more uncertain: the median $B$ is $11.7$
to $17.3$ bits, against $3.5$ to $4.3$ bits in infill (Table~\ref{tab:controls}).

\section{Measuring $\C$ on a real model}
\label{sec:protocol}

\subsection{The estimator}
\label{sec:estimator}

The marginal entropies in Eq.~\eqref{eq:tc} come from one forward pass. The joint
entropy $H(Y_S\given c)$ does not, so we build it from the chain rule, which holds for
any order $\pi$ of the positions:
$H(Y_S\given c)=\sum_t H(Y_{\pi(t)}\given Y_{\pi(<t)},c)$. In practice we draw a random
order $\pi$ and reveal the block one position at a time. At step $t$ we read the model's
distribution for position $\pi(t)$ given the context and the tokens revealed so far,
record its \emph{exact} entropy, sample a token from it, and write the token in. The
recorded entropies add up to one estimate of the joint entropy,
\begin{equation}
  \hat H_{\pi}(Y_S\given c) \;=\; \sum_{t=1}^{|S|}
  H_\theta\!\big(Y_{\pi(t)} \given y_{\pi(<t)}, c\big),
  \qquad
  y_{\pi(t)} \sim p_\theta\!\big(\cdot \given y_{\pi(<t)}, c\big),
  \label{eq:meter}
\end{equation}
and we average over $M$ independent orders:
\begin{equation}
  \Chat_M(S\given c) \;=\; \underbrace{\sum_{i\in S} H_\theta(Y_i\given c)}_{B(S\given c)}
  \;-\; \frac{1}{M}\sum_{m=1}^{M}\hat H_{\pi_m}(Y_S\given c).
  \label{eq:chat}
\end{equation}
If the conditionals compose into one joint, $\hat H_\pi$ is unbiased for $H(Y_S\given c)$
under \emph{any} single order, and averaging only reduces the variance. If they do not,
$\Chat$ is still well defined, but its meaning changes. We postpone that case to
\S\ref{sec:meaning}.

We use $M{=}64$ throughout. For each block we report the standard error
$\mathrm{sd}_m(\hat H_{\pi_m})/\sqrt{M}$, and we correct for it wherever $\Chat$ enters a
variance decomposition. The estimator touches the model only through its conditionals, so the same code
runs on a checkpoint and on an exact oracle. On synthetic tasks whose $\C$ is known in
closed form (App.~\ref{app:oracle}) it recovers the analytic value to within $0.0087$
bits.

\subsection{Validating the readout}
\label{sec:bugimpact}

Everything we report rests on one contract: given a masked sequence and a position $p$,
return the model's distribution for the token at $p$. Three gates test that contract on
every checkpoint before any number is produced.

\begin{figure}[t]
\centering
\includegraphics[width=0.86\textwidth]{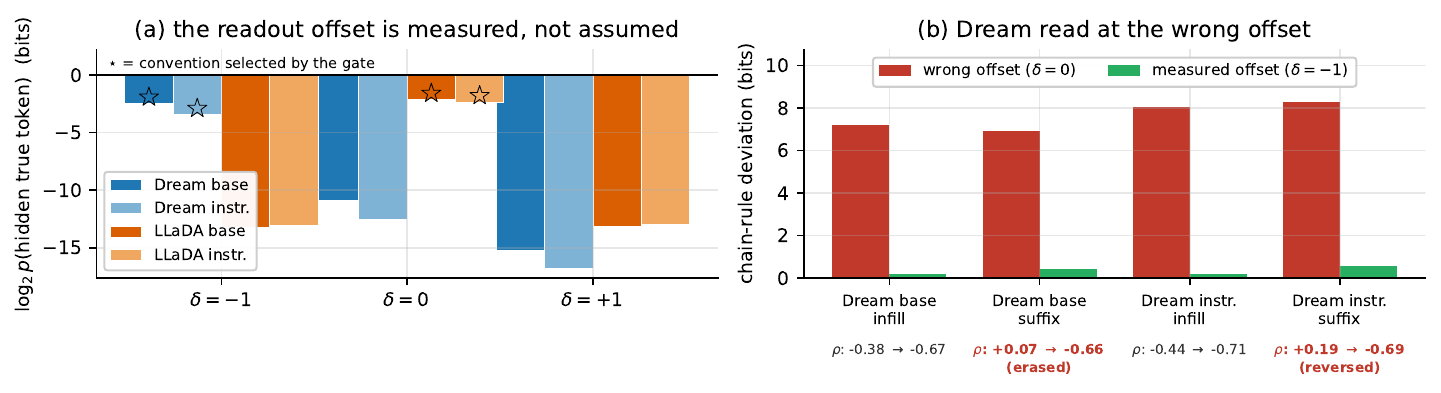}
\caption{\textbf{(a)} The offset gate: log-probability of the hidden true tokens, read
at index $p+\delta$. Stars mark the convention the gate selects: Dream and LLaDA select
\emph{opposite} ones. \textbf{(b)} The
offset ablation, with only the readout changed. Chain-rule deviation inflates $14$ to
$41\times$ on both Dream checkpoints. Under each pair, $\rho$ is the correlation between
confidence and decoding error at the wrong and at the measured offset.}
\label{fig:gates}
\end{figure}

\paragraph{Gate 1: which logit index carries position $p$?} We mask
a block and score the log-probability the model gives to the tokens that were
\emph{actually} there, reading at index $p+\delta$ for $\delta\in\{-1,0,+1\}$. The true
tokens are hidden, so the model cannot win by copying, and only the correct $\delta$
scores well. The winner must be the same for $k{=}1$ and $k{=}4$ and must beat the
runner-up by at least one bit. Without being told, the gate recovers each model's
documented convention, by margins of $6.9$ to $11.0$ bits (Fig.~\ref{fig:gates}a,
Table~\ref{tab:gates}).

\paragraph{Gate 2: is the forward pass reproducible?} Two identical calls must return
identical logits. In addition, the two batch shapes that the estimator combines must
agree on every argmax and on $B$ to within $0.074$ bits RMS, a tenth of the smallest
error we report. All four checkpoints pass it on a second GPU, and the main run used a weaker form
(App.~\ref{app:gates}). We also check a \emph{structural invariant} on every block: the
first position decoded sequentially is drawn from the same distribution as the parallel
commit, so its argmax must never disagree with it.

\paragraph{Gate 3: does it write?} We greedy-decode into a fully masked sequence at the
calibrated offset and read the text. At a wrong offset the decode collapses on all three
checkpoints we probed: one token fills $21$ to $32$ of $32$ prose positions, against
runs of at most $3$ at the calibrated offset (App.~\ref{app:probe}).

\paragraph{What a wrong offset does.} To see what the gates protect against, we rerun
Dream with the offset changed from the measured $-1$ to the naive $0$. The corpus, the
blocks, the reveal orders and the code path stay fixed. Three quantities move
(Fig.~\ref{fig:gates}b).

\emph{An inconsistency diagnostic is inflated.} The chain-rule deviation
\cite{chainruledev} takes one fixed assignment of the block and measures how much its
chain-rule log-probability changes with the reveal order. It is exactly zero when the
conditionals compose into one joint. At the wrong offset it reads $6.9$ to $8.3$ bits,
against $0.18$ to $0.59$ at the right one, a factor of $14$ to $41$. A reader would
conclude that Dream's conditionals are far from any joint.

\emph{A headline correlation disappears.} A common outcome variable is the
\emph{decoding error}: how often a parallel commit disagrees with a sequential
reference decode. At the right offset its rank correlation with confidence is $-0.66$
and $-0.69$ in the suffix regime. At the wrong offset it is $+0.07$ $[-0.06,0.19]$ on Dream-base and $+0.19$ $[0.07,0.31]$ on Dream-instruct. The wrong offset erases the
correlation on one checkpoint and reverses it on the other. A reader would conclude that confidence says nothing about decoding error.

\emph{$\Chat$ becomes impossible.} In the infill regime the median $\Chat$ is $-4.3$
and $-4.8$ bits at the wrong offset, against $+0.78$ and $+0.05$ at the right one. No
joint has $\C<0$, so a strongly negative estimate cannot be a finding. Here the
estimator flags the error that the other two statistics hide.

% Figure 2 is placed here so that it lands on the page where Section 4 starts.
\begin{figure}[t]
\centering
\includegraphics[width=0.86\textwidth]{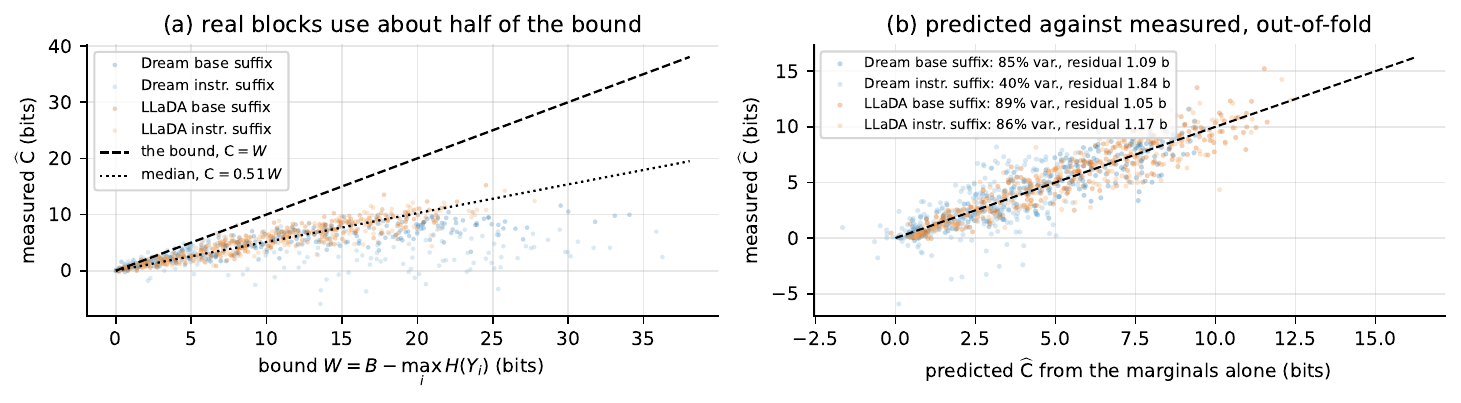}
\caption{\textbf{Suffix regime only.} The infill settings are in
Table~\ref{tab:explains}. \textbf{(a)} Measured $\Chat$ against the bound
$W=B-\max_i H(Y_i)$ of \eqref{eq:width}. The dashed line is $\C=W$ and the dotted line
is the median fraction of $W$ outside Dream-instruct. \textbf{(b)} The best
marginal-only prediction against measured $\Chat$, out-of-fold.}
\label{fig:explains}
\end{figure}

\section{How much of $\C$ the marginals explain}
\label{sec:explains}

\paragraph{Protocol.} We apply the estimator \eqref{eq:chat} to the base and instruct
checkpoints of Dream-Coder 7B \cite{dreamcoder} and LLaDA 8B \cite{llada}, on $247$ to
$516$ blocks per checkpoint and regime (App.~\ref{app:details}). In each of the eight
settings we then ask how well $\Chat$ can be predicted from the marginals of the block
alone. We sweep $k$-NN and ridge over five feature sets built from the per-position
top-$8$ probabilities and entropies, $40$ candidates in all, and keep the best. Every
score is out-of-fold, and all blocks of a document share a fold.

Two definitions make the numbers readable. The \emph{residual} of a predictor is its
out-of-fold root-mean-square error in bits, after the estimator's own variance is
subtracted ($0.7$ to $7.5\%$ of $\mathrm{Var}[\Chat]$). The \emph{reference} is not a fitted model. It is a
reference that knows the marginal entropies of each block, hence its bound $W$, and
nothing about the coupling: it treats $\C$ as uniform on $[0,W]$ and guesses $W/2$. A bound is not an error, and the
reference turns it into one that the residual can be compared with.

Dream-instruct behaves differently from the other three checkpoints. The ranges below
exclude it, and we return to it at the end of the section. On a minority of blocks
$\Chat$ falls outside $[0,W]$. We keep those blocks and discuss them in
\S\ref{sec:meaning}.

\paragraph{The marginals predict $\Chat$ to about one bit.} The best marginal-only
predictor leaves a residual of $0.74$ to $1.17$ bits (Table~\ref{tab:explains},
Fig.~\ref{fig:explains}b). For scale, the median bound $W$ is $1.3$ to $10.8$ bits
across all eight settings. The residual is $1.8$ to $3.7$ times below the reference. A
reference that knew $\C_{\max}$ instead of $W$ would be narrower, so these ratios are on
the generous side.

\paragraph{Most of that is the bound itself.} Fig.~\ref{fig:explains}a shows where real
blocks fall inside $[0,W]$. They do not spread over it. Blocks with $W$ above one bit
sit at a median of $0.44$ to $0.56$ of $W$. A predictor with a single parameter,
$f\cdot W$, therefore already does well. On LLaDA-base it matches the best predictor
($0.74$ and $1.07$ bits against $0.74$ and $1.05$). The full marginal profile helps
most on Dream-base in the suffix regime, where it lowers the residual from $1.81$ to
$1.09$ bits. The six rules add little beyond $W$: given $W$ the rank correlation of any
of the six rules with $\Chat$ lies between $-0.28$ and $+0.23$ (Table~\ref{tab:paired}).

\paragraph{Dream-instruct is the exception, on prose.} Six of the eight settings agree
closely. The two Dream-instruct settings do not ($1.62$ and $1.84$ bits), and the excess
is concentrated on prose.
Under one pooled predictor its residual is $2.23$ and $2.58$ bits on prose, above every
other checkpoint's prose residual, and $0.92$ and $1.28$ bits on code, its training
domain (Table~\ref{tab:domain}). Prose is also where its chain-rule deviation is the
largest we measure, and where the KL cost of a parallel commit exceeds $\Chat$ by
several bits (\S\ref{sec:meaning}). On Dream-instruct prose, then, $\Chat$ does not
measure the cost of a commit, and a poor fit is what one should expect. It is not a generic instruction-tuning effect, since LLaDA-instruct sits with the base
checkpoints.

\paragraph{The result holds at $k{=}8$.} Repeating the measurement with blocks of eight
positions on both base checkpoints gives residuals of $1.60$ to $2.59$ bits. That is
$3.4$ to $5.0$ times below the reference, against $1.8$ to $3.7$ times at $k{=}4$
(Table~\ref{tab:k8}).

\begin{table}[t]
\centering
\small
\setlength{\tabcolsep}{4pt}
\begin{tabular}{lrrrrrrrrr}
\toprule
 & & \multicolumn{2}{c}{measured $\widehat{\mathrm{C}}$} & \multicolumn{2}{c}{\% Var explained} & \multicolumn{3}{c}{residual (bits)} & bound \\
\cmidrule(lr){3-4}\cmidrule(lr){5-6}\cmidrule(lr){7-9}
Model / regime & $n$ & mean & s.d. & $B$ & best & best & $f\cdot W$ & reference & $W$ \\
\midrule
Dream base / infill & $494$ & $1.70$ & $2.25$ & $65$ & $75$ & $1.13$ & $1.41$ & $2.04$ & $2.0$ \\
Dream base / suffix & $247$ & $4.90$ & $2.79$ & $81$ & $85$ & $1.09$ & $1.81$ & $4.06$ & $10.8$ \\
Dream instr. / infill & $494$ & $0.54$ & $1.89$ & $3$ & $26$ & $1.62$ & $2.03$ & $2.53$ & $1.3$ \\
Dream instr. / suffix & $247$ & $2.54$ & $2.37$ & $15$ & $40$ & $1.84$ & $3.59$ & $4.01$ & $6.7$ \\
LLaDA base / infill & $516$ & $1.78$ & $2.38$ & $86$ & $90$ & $0.74$ & $0.74$ & $1.52$ & $1.7$ \\
LLaDA base / suffix & $258$ & $5.47$ & $3.24$ & $88$ & $89$ & $1.05$ & $1.07$ & $3.48$ & $10.3$ \\
LLaDA instr. / infill & $516$ & $1.67$ & $2.28$ & $78$ & $85$ & $0.87$ & $0.95$ & $1.60$ & $1.6$ \\
LLaDA instr. / suffix & $258$ & $5.03$ & $3.13$ & $81$ & $86$ & $1.17$ & $1.34$ & $3.44$ & $9.5$ \\
\bottomrule
\end{tabular}
\caption{\textbf{How well the marginals predict $\widehat{\mathrm{C}}$.}
\emph{measured $\widehat{\mathrm{C}}$}: mean and noise-corrected s.d.\ in bits.
\emph{\% Var explained}: out-of-fold $R^2$ of $B$ alone and of the best of $40$
predictors (\S\ref{sec:explains}). \emph{residual}: noise-corrected out-of-fold error of
that predictor, of the one-parameter $f\cdot W$, and of a reference that knows the
marginal entropies, with $\mathrm{C}$ uniform on $[0,W]$.
$W$: median bound.}
\label{tab:explains}
\end{table}

\section{Comparing the six marginal rules}
\label{sec:fidelity}

We now score the six rules of \S\ref{sec:worstcase} against measured $\Chat$. The score
is the Spearman rank correlation $\rho$ between a rule's safety score and $\Chat$ across
blocks. A good rule calls a block safe when its $\Chat$ is low, so a faithful rule has a
negative $\rho$ (Table~\ref{tab:fidelity}).

\paragraph{Every rule tracks $\Chat$, mostly through certainty.} All $48$ correlations,
six rules in eight settings, are negative and exclude zero under a document-clustered
bootstrap. That much is expected. Since $\C\le W$, any rule that tracks certainty also
tracks the bound, and conditioning on $W$ removes most of the correlation
(\S\ref{sec:explains}). The informative part is how the rules compare with each other.

\paragraph{How a rule pools matters more than what it reads.} In every one of the eight
settings, the pooled rule beats the min rule that reads the same statistic: mean
confidence beats minimum confidence, mean margin beats minimum margin, and summed
entropy beats maximum entropy. The gain is $0.066$ to $0.235$ in $|\rho|$, and a paired
bootstrap excludes zero in all $24$ comparisons (Table~\ref{tab:paired}). Gating on the
least confident position, as threshold rules do \cite{fastdllm,dynamicdllm}, is the
weaker way to pool. The pooled rules reach $\rho=-0.83$ to $-0.94$ outside
Dream-instruct.

\paragraph{A prescription from synthetic tasks does not transfer.}
\label{sec:oracle}
Before turning to real checkpoints we scored the same six rules where $\C$ is known
exactly, on $355$ candidate sets from the synthetic list-operation tasks of ParallelBench
\cite{parallelbench} (App.~\ref{app:oracle}). There the top-2 margin ranks $\C$ far
better than confidence, by $0.394$ $[0.296,0.494]$ in $\rho$ under a paired bootstrap
(Table~\ref{tab:proxy}). On real checkpoints that advantage is gone. Among the min
versions, confidence is better in seven of eight settings, significantly so in six, by
up to $0.189$ $[0.121,0.264]$. Among the pooled rules neither statistic dominates. Mean
confidence is significantly better in five settings, mean margin in the two
Dream-instruct settings, and one is a tie. Likewise, summed entropy beats mean
confidence significantly in three settings and loses significantly in two
(Table~\ref{tab:paired}).

\paragraph{What this does and does not say.} These are statements about predicting
$\Chat$. Whether a pooled rule makes a better decoder is a separate question that needs
an end-to-end comparison. We also do not use the usual outcome variable, decoding error,
as the target. It is a weak stand-in for $\Chat$, with a correlation of only $+0.23$ to $+0.64$, and in the suffix
regime it depends on the reveal order of the reference decode (App.~\ref{app:controls}).

\begin{table}[t]
\centering
\small
\setlength{\tabcolsep}{4pt}
\begin{tabular}{lrrrrrr}
\toprule
Model / regime & conf (min) & conf (mean) & marg (min) & marg (mean) & negH (max) & negH (sum) \\
\midrule
Dream base / infill & $-0.72$$^{*}$ & $-0.83$$^{*}$ & $-0.69$$^{*}$ & $-0.83$$^{*}$ & $-0.77$$^{*}$ & $-0.83$$^{*}$ \\
Dream base / suffix & $-0.79$$^{*}$ & $-0.89$$^{*}$ & $-0.65$$^{*}$ & $-0.86$$^{*}$ & $-0.78$$^{*}$ & $-0.88$$^{*}$ \\
Dream instr. / infill & $-0.23$$^{*}$ & $-0.33$$^{*}$ & $-0.28$$^{*}$ & $-0.37$$^{*}$ & $-0.17$$^{*}$ & $-0.23$$^{*}$ \\
Dream instr. / suffix & $-0.46$$^{*}$ & $-0.54$$^{*}$ & $-0.45$$^{*}$ & $-0.58$$^{*}$ & $-0.36$$^{*}$ & $-0.45$$^{*}$ \\
LLaDA base / infill & $-0.73$$^{*}$ & $-0.86$$^{*}$ & $-0.69$$^{*}$ & $-0.85$$^{*}$ & $-0.79$$^{*}$ & $-0.89$$^{*}$ \\
LLaDA base / suffix & $-0.78$$^{*}$ & $-0.89$$^{*}$ & $-0.60$$^{*}$ & $-0.83$$^{*}$ & $-0.84$$^{*}$ & $-0.94$$^{*}$ \\
LLaDA instr. / infill & $-0.75$$^{*}$ & $-0.85$$^{*}$ & $-0.71$$^{*}$ & $-0.84$$^{*}$ & $-0.80$$^{*}$ & $-0.87$$^{*}$ \\
LLaDA instr. / suffix & $-0.76$$^{*}$ & $-0.89$$^{*}$ & $-0.61$$^{*}$ & $-0.85$$^{*}$ & $-0.81$$^{*}$ & $-0.90$$^{*}$ \\
\bottomrule
\end{tabular}
\caption{\textbf{Every marginal rule against measured $\mathrm{C}$.} Spearman $\rho$
between each rule's safety score and $\widehat{\mathrm{C}}$, so a faithful rule is
negative. $^{*}$: the document-clustered bootstrap interval excludes zero. Paired
comparisons between rules are in Table~\ref{tab:paired}.}
\label{tab:fidelity}
\end{table}

\section{What $\Chat$ does and does not measure}
\label{sec:meaning}

Eq.~\eqref{eq:tc} to \eqref{eq:width}, and the reading of $\C$ as the KL cost of a
parallel commit, assume that the conditionals compose into one joint. A dLLM's any-order
conditionals need not \cite{mlminconsist}. The estimator does not need the assumption:
$\Chat$ is always $B$ minus an order-averaged chain-rule entropy.

\paragraph{Three quantities that agree only for a consistent model.} Let $q_\pi$ be the
joint over $S$ that the model produces when $S$ is revealed in order $\pi$, let
$\bar q=\mathbb{E}_\pi q_\pi$ be the joint of a sampler whose order is uniformly random,
and let $p_i$ be the parallel marginals. Three numbers matter. The first is what the
estimator returns, $\mathbb{E}\,\Chat_M=B-\mathbb{E}_\pi H(q_\pi)$, since each
$\hat H_\pi$ is unbiased for $H(q_\pi)$ whatever the conditionals. The second is the
cost of a parallel commit against a fixed-order sampler, averaged over orders,
$\mathbb{E}_\pi\,\mathrm{KL}(q_\pi\,\|\,\prod_i p_i)$. The third is its cost against the
random-order sampler, $\mathrm{KL}(\bar q\,\|\,\prod_i p_i)$.
The last two are tied by an exact identity,
\begin{equation}
  \mathbb{E}_\pi\,\mathrm{KL}(q_\pi\,\|\,\textstyle\prod_i p_i)
  \;=\;\mathrm{KL}(\bar q\,\|\,\textstyle\prod_i p_i)\;+\;I(\pi;Y_S),
  \label{eq:orderinfo}
\end{equation}
where $I(\pi;Y_S)\le\log_2 k!$ is the information that the reveal order carries about
the outcome. When the conditionals compose, all three numbers equal $\C$ and
$I(\pi;Y_S)=0$. When they do not, there are two cases.

When each reveal order keeps the parallel marginals, $\mathbb{E}\,\Chat_M$ equals the
per-order KL cost, the left side of \eqref{eq:orderinfo}. It is still a cost, and it
exceeds the cost against a random-order sampler by the order information. Two fair bits
show the gap. Suppose one order makes the second bit agree with the first with
probability $0.9$, and the other order makes it disagree with probability $0.9$. Each
order costs $0.531$ bits, while the mixture of the two is uniform and costs nothing.

When the reveal orders do not keep the parallel marginals, $\Chat$ is not a divergence
at all, and it can be negative.

\paragraph{How far $\Chat$ is from the cost of a commit.} We measured both terms of
\eqref{eq:orderinfo} directly on a subset of blocks (App.~\ref{app:probe},
Table~\ref{tab:probe}). On its random blocks, the per-order KL cost exceeds $\Chat$ by a
median of $0.4$ to $1.1$ bits on LLaDA-base and Dream-base prose and in their suffix
regime. $\Chat$ still ranks blocks like the KL cost (Spearman $0.95$ and $0.90$, pooled
across settings). So on these checkpoints $\Chat$ orders blocks by the cost of a commit
but understates its size, and the residuals of \S\ref{sec:explains} are residuals for
$\Chat$, not for the KL cost. On Dream-instruct prose the gap is $4.4$ and $5.6$ bits,
and there $\Chat$ does not measure the cost.

\paragraph{Impossible values.} No joint gives $\C<0$ or $\C>W$, so the blocks where
$\Chat$ leaves $[0,W]$ are a direct sign that something is off. Beyond two standard
errors, $0\%$ to $25\%$ of blocks fall below zero and $0.4\%$ to $9.5\%$ above $W$
(Table~\ref{tab:domain}). Most of these are tiny. Outside Dream-instruct $79\%$ of these
are within $0.2$ bits, $86\%$ of them on code. There the standard error is too small to trust (App.~\ref{app:probe}). The substantive cases, beyond $0.2$ bits, are
$0.4\%$ to $5.9\%$ of blocks outside Dream-instruct and $8.1\%$ and $14.6\%$ on it. We
report these rates rather than clipping. Excluding the blocks below zero moves
Dream-instruct's residual from $1.62$/$1.84$ to $1.19$/$1.57$ bits and leaves the other
checkpoints essentially unchanged.

\section{Related work}
\label{sec:bg}

\paragraph{The cost of parallel commits.} The cost of a parallel commit has been
analysed as a KL floor in $\C$ \cite{parallelbench,huang}, as the dependency a
factorised step drops \cite{dcd}, and through step counts \cite{feng}, inference
schedules \cite{schedules}, convergence rates \cite{intrinsicdep}, generation order
\cite{genorder} and a Gaussian-field model \cite{gaussiantc}. Dataset-level estimates of
$\C$ date to non-autoregressive translation \cite{huang}. ParallelBench
\cite{parallelbench} supplies the synthetic tasks with analytic $\C$ that we reuse, and
finds that a confidence rule loses to random selection there once $\C>0$. We add an
estimate of $\C$ per candidate set on real checkpoints and ask how much of it the
marginals recover.

\paragraph{Dependency-aware decoders.} Many recent decoders account for the dependency
between commits. They estimate it from attention \cite{dapd,geoblock,clad,dos},
dependency graphs \cite{dawn}, learned mutual information \cite{neuralmi} or a learned
predictor \cite{demask}, or they coordinate commits through a mean-field step
\cite{meanfield}, conditional independence tests \cite{punt}, an auxiliary
autoregressive model \cite{locallycoherent}, a learned policy \cite{tracelock} or a
coordination pass \cite{cocommit}. We measure how well the marginals predict the
dependency of a fixed set, not what such a decoder gains.

\paragraph{Inconsistent conditionals.} That the conditionals of a masked language model
need not compose into any joint is known \cite{mlminconsist}, with consequences for
sampling \cite{glauber} and denoising order \cite{pathdenoise}. Our diagnostic, the
chain-rule deviation, is from \cite{chainruledev}.

\section{Limitations}
\label{sec:limitations}

\paragraph{$\Chat$ is not always a total correlation.} Without a single joint there is
no single quantity that deserves the name total correlation, and $\Chat$ can understate
the KL cost of a commit or fail to measure it (\S\ref{sec:meaning}).

\paragraph{What the residual bounds.} Our predictor reads per-position top-$8$
probabilities and entropies. That subsumes every single-step rule we score, at any
aggregation, but not the full marginal and not the cross-step stability rules of
\cite{less,stability}, which read more than one decoding step. A better predictor could
only lower the residual, so we state it as an upper bound, not an impossibility. Choosing the winner over $40$
candidates on the same out-of-fold scores we report biases the residual downward, but
selecting on inner folds and scoring on outer ones moves it by at most $0.035$ bits
(App.~\ref{app:nested}). Quadrupling $M$ to $256$ on LLaDA-base moves it by at most
$0.034$ bits (App.~\ref{app:budget}).

\paragraph{Scope.} $k{=}4$ blocks with $k{=}8$ as a check, greedy reference decodes, one
$128$-token window per document, English prose and Python (App.~\ref{app:details}).
Four checkpoints across two families, run on raw text without chat formatting, which
keeps inputs comparable but is not how an instruct model is deployed. These numbers
describe one decoding step on fixed contiguous blocks, not end-to-end generation
quality. A deployed decoder chooses its sets, may commit larger ones, and may sample
rather than take the argmax. Numbers measured on one GPU type need not reproduce bit for bit on
another (App.~\ref{app:probe}).

\section{Conclusion}

A rule that reads per-position marginals cannot determine the dependency of a parallel
commit, only bound it by $W$. On real checkpoints that bound carries most of the
information. On four checkpoints from two families, the marginals predict the estimate
$\Chat$ with a residual of $0.74$ to $1.17$ bits outside Dream-instruct, $1.8$ to $3.7$
times below a reference that ignores the coupling, mostly because real blocks use a
stable fraction of their bound. How a rule pools over the set matters more than which
statistic it reads. Two cautions apply. $\Chat$ ranks blocks like the KL cost of a
parallel commit but sits below it on prose and in the suffix regime, and on
Dream-instruct prose it does not measure that cost. And we score fixed sets, so what a
dependency-aware decoder gains needs a matched-compute decoding study.

Two widely used open dLLMs require opposite logit-position conventions. The wrong one
collapses greedy prose on both families, yet a pipeline that never decodes text does not
notice. The cheapest defence is to check the readout against something the model cannot
see, and to report the check.

\clearpage
\appendix
% Appendix tables are set where they are discussed when they fit, and otherwise at the
% top of the next page while the text runs on, so no page is left half empty.
\renewcommand{\topfraction}{0.9}\renewcommand{\bottomfraction}{0.8}
\renewcommand{\textfraction}{0.08}\renewcommand{\floatpagefraction}{0.85}
\setcounter{topnumber}{3}\setcounter{bottomnumber}{2}\setcounter{totalnumber}{4}
\makeatletter
\renewenvironment{table}[1][]{\@float{table}[!htb]}{\end@float}
% two related tables stacked in one float
\newcommand{\tablestack}[2]{%
  \@float{table}[!htb]%
  \begingroup\renewenvironment{table}[1][]{\def\@captype{table}}{\par}%
  \input{#1}\vspace{16pt}\input{#2}\endgroup
  \end@float}
\makeatother
\section{Experimental details}
\label{app:details}

\paragraph{Corpus.} $285$ documents: for English prose, $121$ WikiText-103 articles of at
least $600$ characters from \texttt{wikitext-103-raw-v1}, and for Python, all $164$ OpenAI
HumanEval problems. The first $128$ tokens of each document under the
checkpoint's own tokenizer form one window, and shorter documents are dropped, so Dream
sees all $121$ prose documents and $126$ of the code problems ($247$ windows) and LLaDA
sees $121$ and $137$ ($258$ windows).

\paragraph{The two regimes.} A block is $k$ consecutive positions inside a window.
\emph{Suffix} masks everything from the block to the end of the window, as a decoder
continuing a prompt sees it, with one block per window starting at its midpoint.
\emph{Infill} masks only the block, with two blocks per window at random starts from the
window's second quarter onward, spaced $4k$ apart. This gives $247$ or $258$ suffix
blocks and $494$ or $516$ infill blocks per checkpoint.

\paragraph{Estimator and diagnostics.} $M{=}64$ reveal orders for $\Chat$, $8$ orders for
the chain-rule deviation, $8$ reference orders for the order-sensitivity control, and the
top $8$ probabilities per position for the profile features. Reference decodes are
greedy.

\paragraph{Variance correction.} The estimator is noisy, so
$\mathrm{Var}[\Chat]=\mathrm{Var}[\C]+\mathbb{E}[\mathrm{se}^2]$, and the same offset
appears in any out-of-fold squared error. We subtract the measured
$\mathbb{E}[\mathrm{se}^2]$ from both before forming an $R^2$ or a residual.

\paragraph{Predictors.} Five document-grouped folds under seed $0$, so no document is
split across train and test. Features are standardised. The five feature sets are the
minimum confidence, $B$, per-position confidence and entropy, the full profile
(per-position top-$8$ probabilities and entropy, positions sorted by confidence), and the
full profile with both scalars. The sweep is $k$-NN with $k\in\{5,10,20,40\}$, used when
$k<n/6$, and ridge with $\lambda\in\{0.1,1,10,100\}$ and an unpenalised intercept, over
all five sets, which gives $40$ candidates.

\paragraph{Compute.} One A100 80GB. Both families run in \texttt{bfloat16} under
\texttt{transformers} 4.46.2, which Dream's remote code requires. A full measurement run
takes $4{,}545$ batched forward calls per Dream checkpoint and $4{,}762$ per LLaDA
checkpoint. Everything in
\S\ref{sec:worstcase} and App.~\ref{app:oracle} is exact and runs on CPU.

\paragraph{Availability.} Code, configurations, per-block results and the verification probe are at
\url{https://github.com/DovudAsadov/dllm-total-correlation}, tag \texttt{paper-numbers}.

\section{Readout gates}
\label{app:gates}
Table~\ref{tab:gates} reports Gate 1 for every checkpoint: the log-probability of the
hidden true tokens at each candidate offset, the offset the gate selects, and its margin
over the runner-up. Every checkpoint selects its documented convention.
\begin{table}[!ht]
\centering
\footnotesize
\setlength{\tabcolsep}{3pt}
\begin{tabular}{lccrrrrccc}
\toprule
 & \multicolumn{2}{c}{readout offset} & \multicolumn{3}{c}{$\log_2 p(\text{hidden true token})$ read at $p{+}\delta$} & \multicolumn{2}{c}{gap (bits)} & & \\
\cmidrule(lr){2-3}\cmidrule(lr){4-6}\cmidrule(lr){7-8}
Model & measured & documented & $\delta{=}{-}1$ & $\delta{=}0$ & $\delta{=}{+}1$ & $k{=}4$ & $k{=}1$ & flips & gates \\
\midrule
Dream-Coder-7B (base) & $-1$ & $-1$ (AR shift) & $-2.47$ & $-10.87$ & $-15.24$ & $8.4$ & $6.9$ & 0 & \checkmark \\
Dream-Coder-7B (instruct) & $-1$ & $-1$ (AR shift) & $-3.44$ & $-12.55$ & $-16.80$ & $9.1$ & $8.4$ & 0 & \checkmark \\
LLaDA-8B (base) & $0$ & $0$ (in place) & $-13.19$ & $-2.13$ & $-13.15$ & $11.0$ & $16.0$ & 0 & \checkmark \\
LLaDA-8B (instruct) & $0$ & $0$ (in place) & $-13.03$ & $-2.34$ & $-12.97$ & $10.6$ & $14.7$ & 0 & \checkmark \\
\bottomrule
\end{tabular}
\caption{\textbf{Gate 1: the readout offset is measured, not assumed.} Log-probability of
the hidden true tokens of a masked $k{=}4$ block, read at index $p{+}\delta$. \emph{gap}: margin of the winning $\delta$ over the
runner-up, at $k{=}4$ and with one hidden token. \emph{flips}: argmax changes between two
identical forward calls.}
\label{tab:gates}
\end{table}

\paragraph{Gate 2 in the main run and in the release.} The main run required two
identical calls to return identical logits, and required that scoring a sequence alone
or inside a batch move no argmax. Both held on all four checkpoints: two identical calls
differed by exactly zero, and batch composition moved no argmax, although it moves
log-probabilities by up to $0.56$ ($0.162$, $0.556$, $0.281$ and $0.224$ on Dream-base,
Dream-instruct, LLaDA-base and LLaDA-instruct). The released gate is stricter: the two
batch shapes the estimator combines must agree on every argmax and on $B$ to within
$0.074$ bits RMS. On a second GPU all four checkpoints pass with the two shapes agreeing
exactly. Scoring a sequence alone, which the measurement never does, flips $1$ of $64$
argmaxes on each Dream checkpoint (App.~\ref{app:probe}).

\paragraph{Gate 3 in the main run and in the release.} The main run only required a
non-empty decode, and every checkpoint wrote fluent text at its calibrated offset. The
Dream-base decode, for example, begins ``Theatre . In 2002 , he appeared in the''. The
released gate also requires a wrong offset to produce the collapse of
\S\ref{sec:bugimpact}, and fails on any error.

\section{Exact results: the worst case and the synthetic audit}
\label{app:exact}

\paragraph{The worst case in detail.}
\label{app:fibre}
The set of joints with a fixed marginal profile (a transportation polytope) is convex
and compact. It always contains the product measure, so $\C=0$ is attainable. At the
other end, a minimum-entropy coupling gives $\C_{\max}=B-\min_P H(P)$, and
Eq.~\eqref{eq:width} follows from $H(P)\ge\max_i H(Y_i)\ge B/k$. When all marginals are
equal, the joint that sets every position equal gives $\C=W=(k{-}1)H(Y_1)$, so the bound
is attained. Because $\C$ is continuous on a connected set, every value between $0$ and
$\C_{\max}$ is realised. The entropy $H$ is concave, so its minimum over the polytope is
attained at a vertex, and for small $k$ we compute $\C_{\max}$ by exact vertex
enumeration: $k$ fair bits span exactly $[0,k{-}1]$ bits. For example, four binary
positions that each have confidence $0.9$ carry $B=1.876$ bits and allow
$\C_{\max}=1.407$ bits.

\paragraph{The exact audit.}
\label{app:oracle}
To score the rules where $\C$ is known exactly we use the synthetic list-operation
family of ParallelBench \cite{parallelbench} and its analytic $\C$. Enumerated and
closed-form $\C$ agree to $1.6\times10^{-14}$ bits. Table~\ref{tab:proxy} gives the six
marginal rules scored against it. Over the $355$ candidate sets the six rules collapse
to four distinct rankings. Every rule also attains a perfect safe-set AUROC, but only
because $\C=0$ holds on this family \emph{if and only if} $B=0$, so any certainty
detector scores perfectly. We therefore treat the ranking as a hypothesis to test on
real checkpoints, and \S\ref{sec:fidelity} finds that it does not hold there.
\begin{table}[!ht]
\centering
\footnotesize
\setlength{\tabcolsep}{4pt}
\begin{tabular}{lrrrr}
\toprule
Rule & Spearman$(\cdot,\mathrm{C})$ [95\% CI] & Kendall & AUROC$_{\mathrm{safe}}$ & gap \\
\midrule
confidence (min), neg-entropy (max) & $-0.203\ [-0.345,\, -0.055]$ & $-0.104$ & $1.0000$ & $0$ \\
confidence (mean) & $-0.201\ [-0.344,\, -0.054]$ & $-0.103$ & $1.0000$ & $0$ \\
margin (min/mean) & $-0.596\ [-0.683,\, -0.499]$ & $-0.539$ & $1.0000$ & $0$ \\
neg-entropy (sum) & $-0.335\ [-0.471,\, -0.185]$ & $-0.198$ & $1.0000$ & $0$ \\
\bottomrule
\end{tabular}
\caption{\textbf{The exact audit.} Six marginal rules against analytic $\mathrm{C}$ on $355$ candidate sets from \textsc{replace\_random} and \textsc{shuffle}, one row per distinct ranking. The safe-set AUROC is perfect because $\mathrm{C}=0$ here exactly when $B=0$. \emph{gap}: score difference on the matched-marginal pair of \S\ref{sec:worstcase}, which differ in $\mathrm{C}$ by $1.42$ bits.}
\label{tab:proxy}
\end{table}

\section{Robustness of the residual}
\label{app:robust}
We test the residuals of \S\ref{sec:explains} against three choices: the estimator
budget $M$, the block size $k$, and the way the predictor is selected. None of them
changes the conclusions.

\paragraph{Estimator budget.}
\label{app:budget}
Table~\ref{tab:meter} traces the estimator's convergence in $M$ on $48$ suffix-regime
blocks per base checkpoint. The standard error falls as $1/\sqrt{M}$.
Table~\ref{tab:msens} repeats the analysis at $M{=}256$ on the main run's blocks, and
the residual moves by at most $0.034$ bits.
\begin{table}[!ht]
\centering
\footnotesize
\setlength{\tabcolsep}{5pt}
\begin{tabular}{rrrrrr}
\toprule
$M$ & s.e.\ (bits) & s.e.$\times\sqrt{M}$ & mean $|\Chat_M-\Chat_{256}|$ & max & fwd/block \\
\midrule
\multicolumn{6}{l}{\emph{Dream base, suffix regime, $n=48$ blocks, $k=4$}} \\
$4$ & $2.821$ & $5.64$ & $2.00$ & $10.65$ & $16$ \\
$8$ & $1.856$ & $5.25$ & $1.34$ & $4.05$ & $32$ \\
$16$ & $1.323$ & $5.29$ & $1.05$ & $3.82$ & $64$ \\
$32$ & $0.882$ & $4.99$ & $0.67$ & $1.67$ & $128$ \\
$64$ & $0.671$ & $5.36$ & $0.39$ & $1.45$ & $256$ \\
$128$ & $0.479$ & $5.42$ & $0.22$ & $0.57$ & $512$ \\
$256$ & $0.348$ & $5.56$ & $0$ & $0$ & $1024$ \\
\midrule
\multicolumn{6}{l}{\emph{LLaDA base, suffix regime, $n=48$ blocks, $k=4$}} \\
$4$ & $1.551$ & $3.10$ & $1.25$ & $7.63$ & $16$ \\
$8$ & $1.121$ & $3.17$ & $0.98$ & $4.46$ & $32$ \\
$16$ & $0.783$ & $3.13$ & $0.50$ & $2.09$ & $64$ \\
$32$ & $0.586$ & $3.31$ & $0.44$ & $1.49$ & $128$ \\
$64$ & $0.432$ & $3.45$ & $0.31$ & $1.15$ & $256$ \\
$128$ & $0.302$ & $3.41$ & $0.14$ & $0.64$ & $512$ \\
$256$ & $0.208$ & $3.33$ & $0$ & $0$ & $1024$ \\
\bottomrule
\end{tabular}
\caption{\textbf{The estimator converges as $1/\sqrt{M}$.} Per-block standard error of
$\Chat_M$ and its deviation from $\Chat_{256}$ on $48$ fixed blocks per checkpoint, with
the forward-pass cost. Each smaller $M$ reuses the first $M$ of the same $256$ orders, so
the deviation understates an independent run's error by $\sqrt{256/(256-M)}$. The main
runs use $M{=}64$.}\label{tab:meter}
\end{table}

\paragraph{Block size.}
\label{app:k8}
Table~\ref{tab:k8} gives the per-setting numbers that \S\ref{sec:explains} quotes as
ranges, for blocks of eight positions on the two base checkpoints.

\paragraph{Selecting the predictor without a winner's curse.}
\label{app:nested}
Table~\ref{tab:nested} repeats the analysis of \S\ref{sec:explains} with the model and
feature-set selection nested inside its own folds. The residual rises by at most
$0.035$ bits.
\begin{table}[!ht]
\centering
\footnotesize
\setlength{\tabcolsep}{5pt}
\begin{tabular}{lrrrrrrr}
\toprule
 & & \multicolumn{1}{c}{$\Chat$} & \% Var & \multicolumn{2}{c}{residual (bits)} & bound & reference \\
\cmidrule(lr){5-6}
Model / regime & $n$ & mean & expl. & $k{=}8$ & $k{=}4$ & $W$ & (bits) \\
\midrule
Dream base / infill & $250$ & $8.36$ & $90$ & $2.19$ & $1.13$ & $14.3$ & $7.37$ \\
Dream base / suffix & $247$ & $15.97$ & $83$ & $2.59$ & $1.09$ & $40.3$ & $12.09$ \\
LLaDA base / infill & $250$ & $9.31$ & $96$ & $1.60$ & $0.74$ & $14.5$ & $6.12$ \\
LLaDA base / suffix & $250$ & $18.29$ & $92$ & $2.14$ & $1.05$ & $36.2$ & $10.64$ \\
\bottomrule
\end{tabular}
\caption{\textbf{The $k{=}8$ replication, per setting.} The protocol of
Table~\ref{tab:explains} with blocks of eight positions on the two base checkpoints, next
to the $k{=}4$ residual. $W$ and \emph{reference} as in Table~\ref{tab:explains}.}\label{tab:k8}
\end{table}

\section{Further results: domains, paired comparisons and controls}
\label{app:corrections}

\paragraph{Residual by domain.} Table~\ref{tab:domain} splits the residual by domain and
reports the blocks whose $\Chat$ falls outside $[0,W]$. Prose is harder to predict than
code on every checkpoint, and Dream-instruct prose stands apart.
\begin{table}[!ht]
\centering
\footnotesize
\setlength{\tabcolsep}{3.5pt}
\begin{tabular}{lrrrrrrr}
\toprule
 & \multicolumn{2}{c}{residual (bits)} & \multicolumn{2}{c}{chain-rule dev.} & \multicolumn{3}{c}{impossible (\% of blocks)} \\
\cmidrule(lr){2-3}\cmidrule(lr){4-5}\cmidrule(lr){6-8}
Model / regime & prose & code & prose & code & $\widehat{\mathrm{C}}<0$ & $\widehat{\mathrm{C}}>W$ & subst. \\
\midrule
Dream base / infill & $1.35$ \scriptsize$[1.17,1.54]$ & $1.01$ \scriptsize$[0.74,1.27]$ & $0.42$ & $0.06$ & $9.5$ & $7.9$ & $5.9$ \\
Dream base / suffix & $1.33$ \scriptsize$[1.12,1.54]$ & $1.01$ \scriptsize$[0.86,1.15]$ & $0.76$ & $0.28$ & $0.0$ & $1.6$ & $1.2$ \\
Dream instr. / infill & $2.23$ \scriptsize$[1.95,2.49]$ & $0.92$ \scriptsize$[0.54,1.35]$ & $0.59$ & $0.03$ & $25.1$ & $8.9$ & $14.6$ \\
Dream instr. / suffix & $2.58$ \scriptsize$[2.21,2.92]$ & $1.28$ \scriptsize$[1.06,1.47]$ & $1.17$ & $0.30$ & $4.9$ & $8.1$ & $8.1$ \\
LLaDA base / infill & $1.00$ \scriptsize$[0.84,1.16]$ & $0.54$ \scriptsize$[0.44,0.65]$ & $0.28$ & $0.04$ & $8.5$ & $4.3$ & $1.4$ \\
LLaDA base / suffix & $1.18$ \scriptsize$[0.97,1.38]$ & $0.93$ \scriptsize$[0.83,1.05]$ & $0.47$ & $0.21$ & $0.0$ & $0.4$ & $0.4$ \\
LLaDA instr. / infill & $1.16$ \scriptsize$[0.98,1.35]$ & $0.58$ \scriptsize$[0.47,0.68]$ & $0.27$ & $0.04$ & $7.6$ & $9.5$ & $2.3$ \\
LLaDA instr. / suffix & $1.45$ \scriptsize$[1.22,1.68]$ & $1.00$ \scriptsize$[0.83,1.17]$ & $0.67$ & $0.20$ & $0.0$ & $0.8$ & $0.4$ \\
\bottomrule
\end{tabular}
\caption{\textbf{Where the residual and the ill-defined blocks are.} \emph{residual}:
noise-corrected residual of one pooled ridge predictor, scored on each domain, with a
document-clustered 95\% interval. \emph{chain-rule dev.}: median, as in \S\ref{sec:bugimpact}.
\emph{impossible}: percentage of blocks more than two standard errors outside $[0,W]$,
and \emph{subst.}\ those beyond $0.2$ bits.}\label{tab:domain}
\end{table}

\paragraph{Paired comparisons.} Table~\ref{tab:paired} gives the paired comparisons
between rules behind \S\ref{sec:fidelity} and the partial correlations given $W$ behind
\S\ref{sec:explains}.
\begin{table}[!ht]
\centering
\footnotesize
\setlength{\tabcolsep}{3.5pt}
\begin{tabular}{lrrrrrr}
\toprule
 & marg$-$conf & marg$-$conf & negH$-$conf & pooled$-$min & \multicolumn{2}{c}{partial $\rho$ given $W$} \\
\cmidrule(lr){6-7}
Model / regime & (min) & (mean) & (sum vs mean) & (upper end) & \multicolumn{2}{c}{over the six rules} \\
\midrule
Dream base / infill & $+0.033$$^{*}$ & $+0.004$ & $-0.000$ & $-0.048$ & $-0.14$ & $+0.04$ \\
Dream base / suffix & $+0.140$$^{*}$ & $+0.037$$^{*}$ & $+0.017$ & $-0.063$ & $-0.28$ & $+0.21$ \\
Dream instr. / infill & $-0.045$$^{*}$ & $-0.036$$^{*}$ & $+0.101$$^{*}$ & $-0.044$ & $-0.21$ & $+0.28$ \\
Dream instr. / suffix & $+0.007$ & $-0.032$$^{*}$ & $+0.094$$^{*}$ & $-0.034$ & $-0.38$ & $+0.30$ \\
LLaDA base / infill & $+0.041$$^{*}$ & $+0.009$$^{*}$ & $-0.027$$^{*}$ & $-0.074$ & $+0.11$ & $+0.23$ \\
LLaDA base / suffix & $+0.189$$^{*}$ & $+0.063$$^{*}$ & $-0.042$$^{*}$ & $-0.068$ & $-0.04$ & $+0.08$ \\
LLaDA instr. / infill & $+0.041$$^{*}$ & $+0.010$$^{*}$ & $-0.017$$^{*}$ & $-0.057$ & $-0.05$ & $+0.06$ \\
LLaDA instr. / suffix & $+0.144$$^{*}$ & $+0.046$$^{*}$ & $-0.012$ & $-0.072$ & $-0.18$ & $+0.22$ \\
\bottomrule
\end{tabular}
\caption{\textbf{Paired comparisons, and what the rules see beyond certainty.} First three
columns: paired document-clustered bootstrap of
$\rho(\text{first},\widehat{\mathrm{C}})-\rho(\text{second},\widehat{\mathrm{C}})$,
positive where the second rule is better, $^{*}$ where the interval excludes zero.
\emph{pooled vs min}: largest upper end of the three pooled-minus-min intervals, negative
where pooling wins all three. Last two columns: range over the six rules of the partial
Spearman correlation with $\widehat{\mathrm{C}}$ given $W$.}\label{tab:paired}
\end{table}

\paragraph{Controls.}
\label{app:controls}
Table~\ref{tab:controls} gives the per-setting medians of the diagnostics.
\begin{table}[!ht]
\centering
\footnotesize
\setlength{\tabcolsep}{5pt}
\begin{tabular}{lrrrrrr}
\toprule
Model / regime & med.\ conf. & $B$ & $\widehat{\mathrm{C}}$ & chain-rule dev. & cond.\ sens. & pairwise dis. \\
\midrule
Dream base / infill & $0.538$ & $4.2$ & $0.78$ & $0.18$ & $0.080$ & $0.00$ \\
Dream base / suffix & $0.159$ & $17.3$ & $5.06$ & $0.42$ & $0.358$ & $0.29$ \\
Dream instr. / infill & $0.575$ & $4.3$ & $0.05$ & $0.20$ & $0.068$ & $0.00$ \\
Dream instr. / suffix & $0.262$ & $11.7$ & $2.24$ & $0.59$ & $0.311$ & $0.31$ \\
LLaDA base / infill & $0.565$ & $3.6$ & $0.68$ & $0.11$ & $0.050$ & $0.00$ \\
LLaDA base / suffix & $0.177$ & $16.6$ & $5.50$ & $0.28$ & $0.366$ & $0.21$ \\
LLaDA instr. / infill & $0.597$ & $3.5$ & $0.55$ & $0.10$ & $0.049$ & $0.00$ \\
LLaDA instr. / suffix & $0.205$ & $14.8$ & $5.03$ & $0.37$ & $0.352$ & $0.21$ \\
\bottomrule
\end{tabular}
\caption{\textbf{Controls.} Medians over blocks.
\emph{chain-rule dev.}: as in \S\ref{sec:bugimpact}. \emph{cond.\ sens.}: total-variation
distance between $p(Y_j\given c)$ and $p(Y_j\given y_i,c)$. \emph{pairwise dis.}:
disagreement between the greedy assignments obtained under $8$ reveal orders.}
\label{tab:controls}
\end{table}

\paragraph{Decoding error is a weak stand-in for $\Chat$.} Parallel-commit error is
usually measured as the disagreement between a parallel commit and a sequential
reference decode. In the suffix regime that variable is not well defined: a greedy
reference decoded under $8$ different reveal orders yields a median of $2$ distinct
assignments, with pairwise disagreement $0.21$ to $0.31$. In the infill regime it is
order-invariant.

\paragraph{Other controls.} Every checkpoint reacts to what is filled in, with a median
total-variation distance of $0.31$ to $0.37$ in the suffix regime and $0.05$ to $0.08$
in infill. The structural invariant of \S\ref{sec:bugimpact} holds on all $3{,}030$
blocks.

\section{Verification on a second GPU}
\label{app:probe}
We reran the estimator with \texttt{scripts/verify/probe.py} on an NVIDIA GB10, on a
subset of the main run's blocks, to measure both terms of \eqref{eq:orderinfo}
(Table~\ref{tab:probe}) and to test the gates where the main run did not. The gaps quoted
in \S\ref{sec:meaning} use the random blocks ($6$ to $13$ blocks per cell), with the
Spearman correlations pooled across settings, and the median order information there is
at most $0.98$ bits.
\begin{table}[!ht]
\centering
\footnotesize
\setlength{\tabcolsep}{4pt}
\begin{tabular}{lrrrrrrrr}
\toprule
 & \multicolumn{5}{c}{all probed blocks} & \multicolumn{3}{c}{random blocks only} \\
\cmidrule(lr){2-6}\cmidrule(lr){7-9}
Model / regime / domain & $n$ & $\widehat{\mathrm{C}}$ & KL & $I(\pi;Y_S)$ & KL$-I$ & $n$ & KL$-\widehat{\mathrm{C}}$ & $I$ \\
\midrule
Dream base / infill / prose & $12$ & $+0.55$ & $+1.24$ & $0.06$ & $+1.03$ & $10$ & $+0.52$ & $0.06$ \\
Dream base / infill / code & $28$ & $+0.00$ & $+0.11$ & $0.01$ & $-0.02$ & $6$ & $-0.04$ & $0.00$ \\
Dream base / suffix / prose & $9$ & $+7.60$ & $+9.14$ & $0.98$ & $+7.75$ & $9$ & $+1.01$ & $0.98$ \\
Dream base / suffix / code & $11$ & $+1.91$ & $+1.35$ & $0.23$ & $+1.13$ & $7$ & $+0.41$ & $0.38$ \\
Dream instr. / infill / prose & $18$ & $+0.14$ & $+5.87$ & $0.88$ & $+5.05$ & $12$ & $+4.45$ & $0.97$ \\
Dream instr. / infill / code & $22$ & $-0.00$ & $+0.01$ & $0.01$ & $-0.00$ & $4$ & $+1.47$ & $0.16$ \\
Dream instr. / suffix / prose & $11$ & $-0.72$ & $+8.44$ & $1.10$ & $+6.95$ & $6$ & $+5.56$ & $0.84$ \\
Dream instr. / suffix / code & $29$ & $+1.18$ & $+1.42$ & $0.14$ & $+1.28$ & $10$ & $+0.18$ & $0.16$ \\
LLaDA base / infill / prose & $15$ & $+1.99$ & $+2.40$ & $0.18$ & $+2.06$ & $13$ & $+0.47$ & $0.28$ \\
LLaDA base / infill / code & $25$ & $-0.00$ & $-0.00$ & $0.00$ & $-0.02$ & $3$ & $-0.31$ & $0.01$ \\
LLaDA base / suffix / prose & $11$ & $+6.77$ & $+7.91$ & $0.37$ & $+7.77$ & $10$ & $+1.09$ & $0.39$ \\
LLaDA base / suffix / code & $6$ & $+3.67$ & $+3.88$ & $0.15$ & $+3.96$ & $6$ & $+1.15$ & $0.15$ \\
\bottomrule
\end{tabular}
\caption{\textbf{Both terms of \eqref{eq:orderinfo}, measured on a subset.} Medians in
bits on an NVIDIA GB10, over up to $24$ flagged and $16$ random blocks per model and
regime (App.~\ref{app:probe}). $\widehat{\mathrm{C}}$ takes $B$ from the estimator's own
forward pass. \emph{KL} is $\mathbb{E}_\pi\mathrm{KL}(q_\pi\,\|\,\prod_i p_i)$, $I$ the
order information, and KL$-I$ the cost relative to a random-order sampler. Where the
conditionals compose, $\widehat{\mathrm{C}}$, KL and KL$-I$ agree and $I=0$.}\label{tab:probe}
\end{table}

On HumanEval windows the offset gate selects the documented convention by $12.6$ to
$15.9$ bits, against $6.9$ to $11.0$ on prose. Gate 2 with its entropy tolerance
(\texttt{scripts/verify/gate2.py}, $16$ prose and code windows in both regimes) passes on
all four checkpoints, and scoring a sequence alone, which the measurement never does,
moves Dream-base by up to $11.3$ bits of entropy at one position. Taking $B$ from the
estimator's own first forward pass instead of the batch of eight changes $2$ of the
$197$ impossibility flags, and $80$ of the $101$ blocks the main run flagged as outside
$[0,W]$ are outside it again ($39$ of $53$ outside Dream-instruct). On nearly deterministic blocks the per-block standard error under-covers: the sampled
paths almost coincide, so it is close to zero and misses the rare branches. Even the KL
cost, which cannot be negative in expectation, shows it. On the near-deterministic
blocks flagged below zero, the KL version is below minus two standard errors on $16$ of $38$, by up to $0.23$ bits.

\paragraph{Reproducibility across GPUs.} On identical blocks $B$ differs between the
A100 and the GB10 by a mean of $0.04$ to $0.19$ bits per setting and by up to $3.5$ bits
on single Dream blocks.

\end{document}